\documentclass[preprint,12pt]{elsarticle}

\usepackage{amssymb}
\usepackage{amsmath}

\usepackage{natbib}
\usepackage{ragged2e}

\usepackage{xcolor}
\usepackage{multirow}
\usepackage{booktabs}
\usepackage[table,xcdraw]{xcolor}

\usepackage{epsfig}
\usepackage{epstopdf}

\usepackage{float}
\usepackage{caption}
\usepackage{subcaption}
\usepackage{graphicx}
\usepackage{hyperref}
\usepackage{cleveref}
\crefformat{figure}{#2Fig.~#1#3}
\crefformat{table}{#2Table~#1#3}
\crefformat{equation}{#2Eq.~#1#3}
\crefformat{section}{#2Section~#1#3}
\hypersetup{colorlinks=true, linkcolor=cyan, filecolor=magenta, urlcolor=cyan, citecolor=cyan}

\usepackage{natbib}

\usepackage{changes}

\usepackage{titlesec}
\titlelabel{\thetitle\quad}

\journal{Applied Soft Computing}

\begin{document}
	\begin{frontmatter}
	    \title{Skeleton-based Robust Registration Framework for Corrupted 3D Point Clouds}
	    \author[label1]{Yongqiang Wang}
	    \author[label1]{Weigang Li\corref{cor1}}
	    \cortext[cor1]{Corresponding author}
	    \ead{liweigang.luck@foxmail.com}
	    \author[label2]{Wenping Liu}
	    \author[label1]{Zhiqiang Tian}
	    \author[label1]{Jinling Li}

	    \address[label1]{Engineering Research Center for Metallurgical Automation and Measurement Technology Ministry of Education, Wuhan University of Science and Technology, Wuhan 430081, China}
	    \address[label2]{School of Information Management and Institute of Big Data and Digital Economy, Hubei University of Economics, Wuhan, 430205, China}
    \begin{abstract}
    	Point cloud registration is fundamental in 3D vision applications, including autonomous driving, robotics, and medical imaging, where precise alignment of multiple point clouds is essential for accurate environment reconstruction. However, real-world point clouds are often affected by sensor limitations, environmental noise, and preprocessing errors, making registration challenging due to density distortions, noise contamination, and geometric deformations. Existing registration methods rely on direct point matching or surface feature extraction, which are highly susceptible to these corruptions and lead to reduced alignment accuracy. To address these challenges, a skeleton-based robust registration framework is presented, which introduces a corruption-resilient skeletal representation to improve registration robustness and accuracy. The framework integrates skeletal structures into the registration process and combines the transformations obtained from both the corrupted point cloud alignment and its skeleton alignment to achieve optimal registration. In addition, a distribution distance loss function is designed to enforce the consistency between the source and target skeletons, which significantly improves the registration performance. This framework ensures that the alignment considers both the original local geometric features and the global stability of the skeleton structure, resulting in robust and accurate registration results. Experimental evaluations on diverse corrupted datasets demonstrate that SRRF consistently outperforms state-of-the-art registration methods across various corruption scenarios, including density distortions, noise contamination, and geometric deformations. The results confirm the robustness of SRRF in handling corrupted point clouds, making it a potential approach for 3D perception tasks in real-world scenarios.
    \end{abstract}

    \end{frontmatter}

	\section{Introduction}
	\label{sec:Introduction}
	3D point cloud data has become a crucial representation in various fields, including autonomous driving, robotics, medical imaging, and augmented reality, due to its ability to capture rich spatial structures with high precision \cite{xu2023sps, stilla2023change}. In these applications, point cloud registration is of critical importance, which aims to align multiple point clouds into a common coordinate system \cite{ma2022effective}. Accurate registration is essential for creating a coherent and complete representation of complex environments, enabling precise localization, mapping, and object recognition \cite{wang2023roreg, wei2023point}. The effectiveness of registration directly influences the performance of downstream applications, making it a key technology in computer vision and 3D perception \cite{yu2024riga}.

	However, real-world point cloud data is often corrupted, introducing substantial challenges to the registration process. Corruptions can stem from sensor limitations, environmental conditions, occlusions, and data pre-processing errors, significantly degrading the quality and structural integrity of point clouds \cite{zhao2024multi}. These corruptions can be broadly categorized into density variations, noise contamination, and transformation distortions. Density variations result in uneven point distributions due to sensor inconsistencies or occlusions, leading to regions with excessive or sparse points. Noise contamination introduces random perturbations that distort point locations, making correspondence estimation unreliable. Transformation distortions, such as shearing and non-rigid deformations, alter the global or local structure of the point cloud, further complicating alignment.

	\begin{figure}[htbp]
		\centering
		\includegraphics[width=\linewidth]{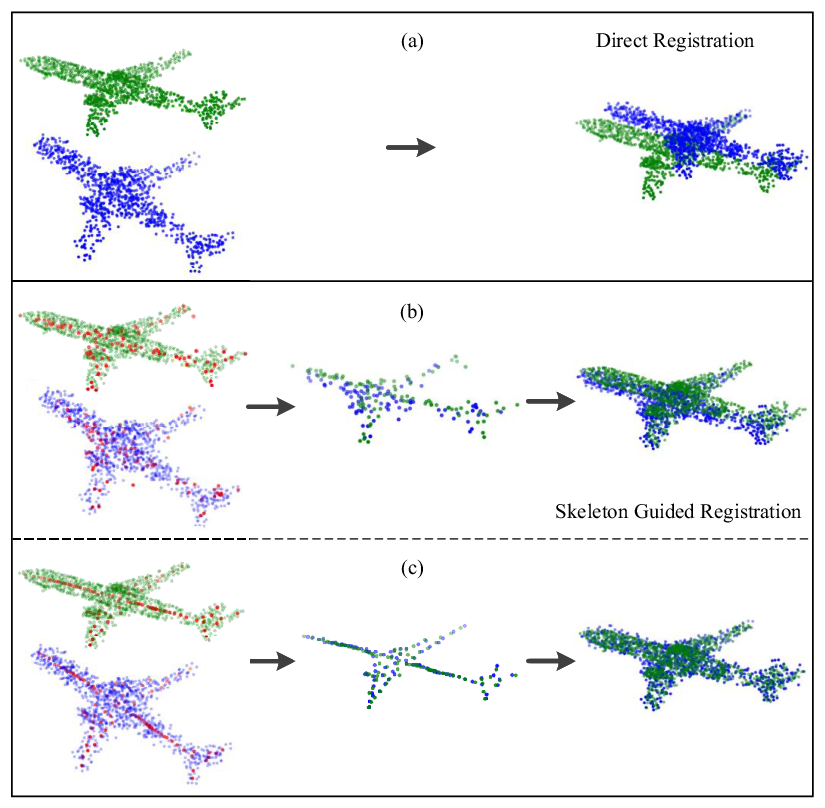}
		\caption{Illustration of corrupted point cloud registration. The target (blue) and source (green) point clouds are affected by Gaussian noise. (a) Registration directly using DCP \cite{wang2019deep}. (b) Skeleton-based registration using Point2Skeleton \cite{lin2021point2skeleton}, showing extracted skeletons (red) and final alignment. (c) The proposed method, which enhances skeleton consistency for improved registration accuracy.}
		\label{fig:ConceptSkeletonRegistration}
	\end{figure}

	Existing registration techniques rely on surface feature extraction or direct point matching, making them highly sensitive to these corruptions \cite{li2024effective,wu2023panet,kang2024equi,fu2021robust}. When point distributions are uneven or distorted, these methods often fail to establish accurate correspondences, leading to misalignment. This limitation highlights the urgent need for a robust registration framework capable of handling corrupted point clouds effectively. As a compact and stable geometric representation, skeleton has shown tremendous potential in various point cloud tasks \cite{wen2023learnable, lin2021point2skeleton}. It can provide a clear and concise abstraction of the point cloud, distilling its most critical components into a reduced set of points \cite{fei2024progressive}. This representation excels in effectively encoding the topological structure of the point cloud, capturing the essential geometric and topological features while maintaining a high degree of stability and simplicity \cite{wen2023learnable}. Therefore, the skeleton points, which represent the core structural features, are less susceptible to noise and distortions commonly found in raw point cloud data, and thus can be used for designing a robust registration scheme of point clouds. From the current, there has been no dedicated research well-tailored for the registration of corrupted point clouds.

	To overcome the challenges posed by corrupted point clouds, A Skeleton-based Robust Registration Framework (SRRF) is presented. This framework enhances registration accuracy and robustness by leveraging the intrinsic skeletal structure of point clouds. The core idea of SRRF is to obtain the optimal alignment by integrating the skeleton representation into the registration process, further integrating the transformations obtained from the alignment of the corrupted point cloud and its skeleton alignment. This method ensures that the alignment takes into account both the detailed local geometric features and the global stability of the skeleton structure, resulting in a robust and accurate registration. Unlike conventional registration methods that rely solely on raw point distributions, as shown in \cref{fig:ConceptSkeletonRegistration}, SRRF significantly enhances the resilience of point cloud registration against corruption-induced misalignments, by integrating skeletal representations and deep feature learning. Experimental evaluations on our benchmark dataset demonstrate that SRRF consistently outperforms existing methods, achieving superior registration accuracy in the presence of density, noise, and transformation distortions. The proposed framework not only provides a robust solution for real-world point cloud registration but also lays the foundation for future advancements in corruption-aware 3D vision applications. In summary, the main contributions of this work are as follows:

	\begin{itemize}
		\item Skeleton-based registration framework: A novel skeleton-based registration strategy for corrupted point clouds is proposed, leveraging the structural stability of skeletons to achieve robust and accurate alignment even under severe corruptions.

		\item Distribution distance loss for skeleton consistency: A distribution distance loss function is proposed to enhance the consistency between source and target skeletons, ensuring that the extracted skeletal structures accurately reflect their respective point clouds and provide a reliable basis for registration.

		\item Comprehensive Benchmarks on corrupted point clouds: Comprehensive experiments are conducted to evaluate the robustness of the current method under various corruption scenarios, including density variations, noise contamination, and transformation perturbations. The experimental results demonstrate that significant performance improvements are achieved by the proposed method over existing approaches and validate the effectiveness of skeleton-based registration under real-world conditions.

	\end{itemize}

	The remainder of this work is structured as follows. In \cref{sec:Related work}, existing approaches related to point cloud registration and skeleton-based representations are systematically reviewed. The problem of point cloud registration is defined in \cref{sec:Problem formulation and dataset construction}, with an emphasis on its challenges in real-world corruption scenarios, and the construction of a corrupted point cloud dataset is described. The SRRF framework is introduced in \cref{sec:Method}, where each module and its functionality are explained in detail.	Experimental evaluations and an analysis of the effectiveness of the current method under various corruption conditions are provided in \cref{sec:Experiments and discussion}. Finally, this work is summarized in \cref{sec:Conclusion}, where the limitations of current methods and potential future research directions are discussed.

    \section{Related work}
    \label{sec:Related work}
    \subsection{Learning-based registration method}
    In recent years, deep learning has achieved significant progress in point cloud registration tasks. Relevant research could be divided into correspondence-based methods and correspondence-free methods \cite{monji2023review}.

    Correspondence-based registration methods establish explicit correspondences between points in the source point cloud and the target point cloud \cite{li2024graph}. These methods dominate the field of point cloud registration \cite{qin2023geotransformer}. The ICP \cite{vizzo2023kiss} algorithm is a foundational correspondence-based registration method that iteratively optimizes the transformation by reducing the distance between matched points. DCP \cite{wang2019deep} approach has built upon the principles of ICP by utilizing depth features to identify corresponding points and employing soft correspondence techniques to achieve point matching. Building upon DCP\cite{wang2019deep}, PRNet \cite{wang2019prnet} introduces self-supervised learning techniques and a keypoint detector to identify keypoint-to-keypoint correspondences in overlapping regions. Recently, Yu et al. \cite{yu2021cofinet} proposed a coarse-to-fine registration framework. In this framework, coarse correspondences are first established through superpoint matching, followed by point-to-point matching within reasonable regions to generate fine correspondences. Building on this approach, Qin et al. \cite{qin2023geotransformer} introduced a geometric Transformer, which learns more representative features by encoding geometric patterns within and between point clouds to achieve robust superpoint matching.

    Correspondence-free registration methods do not directly rely on explicit point-to-point correspondences \cite{wu2023correspondence}. These algorithms involve two key stages \cite{ao2023buffer}. Initially, the global features of the source and target point clouds are encoded using a feature extraction module. Subsequently, the transformation parameters are estimated by assessing the differences between these global features. One of the pioneering learning-based, correspondence-free methods is PointNetLK\cite{aoki2019pointnetlk}, which uses PointNet to obtain global feature descriptors for each point and iteratively applies the Lucas-Kanade algorithm to determine the rigid transformation matrix. Later, PCRNet \cite{sarode2019pcrnet} replaced the LK step with fully connected layers, directly converting the global features into seven-dimensional vectors to represent the rigid transformation. Following this approach, subsequent methods \cite{wu2023correspondence} typically adhere to the embedding regression strategy, with the primary variation being the regression algorithm employed. Unlike correspondence-based methods, these techniques typically learn a pose-sensitive feature in Euclidean space to represent the current pose \cite{2024Towards}. By aligning these features, the pose transformation can be predicted. However, because these methods are trained on discrete point cloud samples, they are sensitive to noise and variations in density \cite{qin2023geotransformer}.

    Real-world point clouds often contain sparse, noisy, and missing data, which pose challenges for traditional registration methods that rely on dense geometric correspondences. While deep learning-based approaches offer robust solutions, they typically require large labeled datasets, which are difficult to obtain due to sensor limitations and occlusions. Recent studies have explored strategies such as knowledge graph-based models \cite{boualaoui2025knowledge}, sparse learning for feature selection \cite{li2020survey}, and deep depth completion \cite{hu2022deep} to address data scarcity issues. Building on these advancements, the SRRF leverages skeletal representations to achieve accurate alignment despite sparse or corrupted data, offering a more resilient alternative to conventional methods.

    \subsection{Point cloud corruption status}
    Point cloud data serves as a fundamental representation for 3D information, extensively utilized in various applications such as autonomous driving, robotics, and augmented reality \cite{zhu2024advancements}. Despite its significance, the practical deployment and application of point cloud data often encounter challenges due to data corruption. Corrupted point cloud data can arise from multiple sources, including sensor noise, environmental interference, and occlusions, which lead to incomplete or inaccurate 3D representations \cite{sun2022benchmarking, ren2022benchmarking}. The corruption of point cloud data significantly impacts the performance of point cloud registration algorithms, which are critical for tasks such as object recognition, mapping, and localization \cite{zhang20233d}. When point cloud data is corrupted, it introduces errors and uncertainties in the registration process, potentially leading to misalignments and incorrect interpretations of the 3D environment \cite{ge2021type}. These inaccuracies can compromise the reliability and safety of systems that rely on precise 3D information \cite{ren2022benchmarking}, such as autonomous vehicles and robotic navigation systems.

    Moreover, corrupted point cloud data poses a challenge for machine learning models trained on clean, ideal datasets \cite{huang2024pointcat}. These models may not generalize well to real-world scenarios where data corruption is prevalent, resulting in degraded performance and robustness \cite{liu2023pcdnf}. Currently, in the field of point cloud classification, the ModelNet-C dataset \cite{sun2022benchmarking, ren2022benchmarking} simulates various types of real-world corruptions to evaluate the performance degradation of different point cloud classification algorithms. This prompts subsequent research to further enhance the robustness of point cloud classification algorithms. However, there has been no work to date that investigates the impact of corrupted point clouds on point cloud registration algorithms, which is a worthy direction for exploration. Addressing this gap requires a thorough investigation of how different types of corruptions affect the accuracy of point cloud registration and proposing methods to mitigate these effects.

    While significant progress has been made in the field of point cloud data for 3D perception, real-world deployments still face numerous challenges. Addressing these challenges requires optimizing sensor design, improving data processing algorithms, and enhancing data quality assessment standards. By addressing these aspects, point cloud data can be better utilized, and further advancements in 3D perception technology can be driven.

    \subsection{Point cloud skeletal representations}
    Point cloud skeletal representations have emerged as a promising approach in the field of computer vision, particularly for tasks involving 3D shape analysis  \cite{pan2023variational} and recognition \cite{wen2023learnable}. The skeletal representation of a point cloud involves abstracting the complex geometric structure of an object into a simplified, yet informative, skeletal form. This abstraction not only reduces the computational complexity but also captures the essential topological and geometrical features of the object \cite{lin2021point2skeleton}.

	Traditional methods employ handcrafted rules to capture geometric attribute relationships and generate skeletal representations. Curve skeletons are widely used for their simplicity and effectiveness in key point extraction \cite{shi2021skeleton}. However, they are specifically defined for tubular geometric shapes, limiting their expressiveness for complex shapes or large-scale point cloud scenes. Medial Axis Transformation (MAT) \cite{wang2022computing} offers a more versatile skeletal representation, capable of encoding arbitrary shapes. Unlike curve skeletons, which are suited for tubular objects, MAT can represent a broader range of geometries. Despite its versatility, MAT is highly sensitive to surface noise, with even minor perturbations causing numerous insignificant spikes. To address this, some approaches \cite{wen2023learnable} have applied simplification techniques to reduce noise distortions, though these methods often suffer from high computational costs and instability. Recently, learning-based approaches \cite{wen2023learnable, lin2021point2skeleton} have leveraged deep neural networks to predict MAT-based skeletons, significantly improving both robustness and computational efficiency. These advanced methods have demonstrated promising results in various 3D vision tasks, such as object recognition and shape reconstruction.

	\section{Problem formulation and dataset construction}
	\label{sec:Problem formulation and dataset construction}

	\subsection{Problem definition of point cloud registration}
	\label{sec:Problem definition of point cloud registration}
	Point cloud registration aims to estimate a rigid transformation $T$ that best aligns the source point cloud $X = \left\{ {{x_1},{x_2}, \cdots ,{x_N}} \right\} \in {\mathbb{R}^{N \times 3}}$ with the target point cloud $Y = \left\{ {{y_1},{y_2}, \cdots ,{y_N}} \right\} \in {\mathbb{R}^{N \times 3}}$, where $N$ is the numbers of points in the respective point clouds. The point clouds $X$ and $Y$ represent two views of the same scene. The transformation $T$ is parameterized by a rotation matrix $R \in \mathrm{SO(3)}$ and a translation vector $t \in {\mathbb{R}^3}$, which can be determined by solving a Procrustes problem. The transformation parameters $R$ and $t$ are obtained by minimizing the least-squares error $E(R,t)$:
	\begin{equation}\label{1}
		E(R,t) = \frac{1}{N}\sum\limits_{i = 1}^N {{{\left\| {R{x_i} + t - {m_{\cal Y}}\left( {{x_i}} \right)} \right\|}^2}}
	\end{equation}
	where ${m_{\cal Y}}\left( {{x_i}} \right)$ maps the corresponding point of ${x_i}$ in $Y$. The centroids of $X$ and $Y$ are defined as:
	\begin{equation}\label{2}
		\bar x = \frac{1}{N}\sum\limits_{i = 1}^N {{x_i}} {\rm{\quad and \quad}}\bar y = 	\frac{1}{N}\sum\limits_{i = 1}^N {{y_i}}
	\end{equation}

	The covariance matrix \( H \) is given by:
	\begin{equation}\label{3}
		H = \sum\limits_{i = 1}^N {{{\left( {{x_i} - \bar x} \right)}^T}} \left( {{y_i} - \bar y} \right)
	\end{equation}

	Using Singular Value Decomposition (SVD), $H$ is decomposed as $H = US{V^T}$. The rotation matrix $R$ and translation vector $t$ that minimize \( E(R,t) \) are computed as:
	\begin{equation}\label{4}
		R = V{U^T}{\rm{\quad and \quad}}t =  - R\bar x + \bar y
	\end{equation}

	\subsection{Challenges in point cloud registration under real-world corruptions}
	While point cloud registration is well-defined in ideal conditions, real-world applications often involve corrupted point clouds due to sensor limitations, environmental noise, and dynamic scene variations. These corruptions can significantly degrade registration performance by altering the geometric structure of the data, leading to misalignment and inaccurate transformations. Density variations cause uneven point distributions, noise contamination introduces random distortions, and transformation perturbations induce structural deformations—all of which challenge traditional and learning-based registration methods.

	To systematically evaluate the robustness of point cloud registration techniques in these challenging scenarios, a comprehensive benchmark dataset is constructed, incorporating diverse corruption patterns. This dataset simulates real-world conditions by introducing controlled distortions, ensuring a rigorous and realistic assessment of registration algorithms. By analyzing performance across various corruption types, more resilient registration techniques capable of handling imperfect and incomplete point cloud data in practical applications are aimed to be developed.

	\subsection{Corruption point cloud dataset construction}
	\label{sec:Corruption point cloud dataset construction}

	In this section, the fundamental design principles\cite{sun2022benchmarking, ren2022benchmarking} for simulating point cloud corruptions that commonly occur in real-world applications are outlined. Specifically, density variations, noise contamination, and transformation perturbations are focused on, as they significantly impact the accuracy and robustness of point cloud registration. These corruptions are introduced due to various factors, such as sensor limitations, environmental interference, and object movement during scanning, posing substantial challenges to point cloud processing and analysis. By adhering to these principles, a controlled and representative dataset is constructed to enable the systematic evaluation of point cloud registration techniques under realistic corruption conditions. In the following sections, the specific corruption types and their impact on registration performance are detailed.

 	\subsubsection{Density variations}
\begin{figure}[!htb]
 		\centering
 		\includegraphics[width=\linewidth]{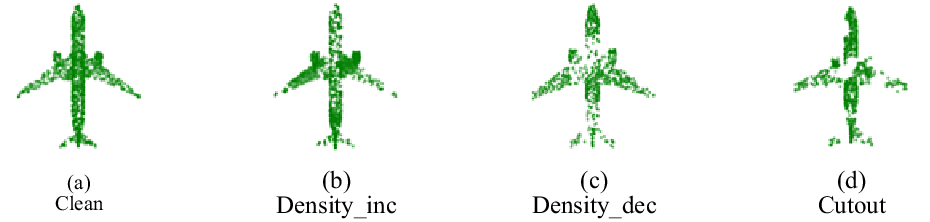}\\
 		\caption{Visualization of density corruptions in point clouds. (a) Clean point cloud. (b) Increased local density due to repeated scanning. (c) Decreased density from occlusions or scanning inconsistencies. (d) Missing regions caused by sensor occlusion or failures.}
 		\label{fig:CorruptionDensity}
 	\end{figure}

 	In real-world scenarios, density variations refer to the uneven distribution of points within a point cloud, which adversely affects data quality and processing reliability. These variations primarily stem from sensor resolution limitations, variations in scanning speed, and inconsistencies in scanning perspective. As a result, local regions may experience increased density, decreased density, or missing data, posing challenges for point cloud analysis. The impact of these corruptions is visually illustrated in \cref{fig:CorruptionDensity}.

 	To systematically evaluate algorithm robustness against density corruption, the following simulation methods are employed:

 	(1) Density\_inc: This corruption occurs when certain regions become abnormally dense due to repeated or high-resolution scanning over a small area. To simulate this effect, regions within the point cloud are randomly selected using the K-Nearest Neighbors (KNN) algorithm to identify point sets, and density is increased by adding new points with random offsets.

 	(2) Density\_dec: This corruption arises when occlusions or poor scanning angles lead to sparsely populated regions. To replicate this scenario, the KNN algorithm is applied to select regions, and a proportion of points is randomly removed to mimic density reduction.

 	(3) Cutout: This corruption results from sensor occlusion or scanning interruptions, causing significant portions of the point cloud to be missing. To simulate this effect, several randomly selected local point clusters are discarded using the KNN algorithm.

 	By incorporating these simulated corruptions into experiments, algorithm robustness can be thoroughly assessed, facilitating the development of methods capable of handling real-world point cloud imperfections.

 	\subsubsection{Noise contamination}
\begin{figure}[ht]
		\centering
		\includegraphics[width=\linewidth]{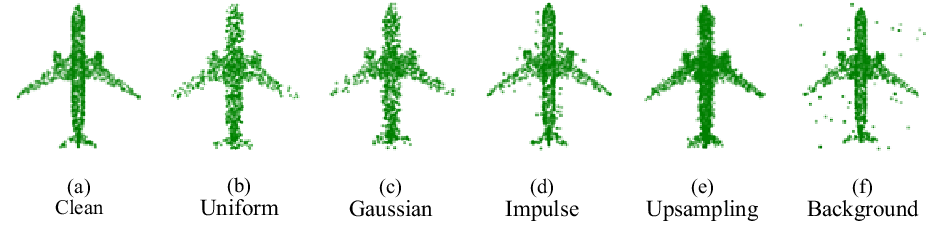}
		\caption{Visualization of noise contamination in point clouds. (a) Clean point cloud. (b) Evenly distributed random noise. (c) Sensor measurement errors following a Gaussian distribution. (d) Sudden signal disturbances causing random point displacements. (e) Extra points introduced due to interpolation errors. (f) Extraneous points simulating real-world scanning clutter.}
		\label{fig:CorruptionNoise}
	\end{figure}

	Noise contamination is a critical issue in point cloud processing, significantly affecting data quality and the accuracy of subsequent applications. Various sources contribute to noise, including inherent errors from acquisition devices due to technological and physical limitations, environmental factors such as lighting and atmospheric conditions, and additional noise introduced during preprocessing, transmission, or storage, such as compression artifacts and digitization errors. These factors lead to five types of noise corruption: Uniform Noise, Gaussian Noise, Impulse Noise, Upsampling Noise, and Background Noise. The effects of these noise types are visually illustrated in \cref{fig:CorruptionNoise}.

	To systematically evaluate algorithm robustness against noise contamination, the following simulation methods are employed:

	(1) Gaussian: This corruption occurs due to sensor measurement errors, where noise follows a Gaussian distribution. To simulate this, Gaussian-distributed random perturbations with a mean of zero are added to each coordinate of the point cloud, constrained within the range of -1 to 1, based on noise severity.

	(2) Uniform: This corruption results from uniformly distributed errors across the dataset. To replicate this effect, noise is randomly sampled from a uniform distribution within a specified range, and the data is normalized accordingly.

	(3) Background: This type of corruption mimics real-world clutter, where extraneous points appear outside the main structure of the point cloud. To simulate this, points are randomly sampled within the bounding cube of the point cloud, with the number of points determined by the severity level.

	(4) Impulse: This corruption simulates sudden signal disturbances, where randomly selected points are significantly displaced. To reproduce this effect, points are selected according to severity, and each is assigned a perturbation with a maximum magnitude of \( \ell_\infty = 0.05 \).

	(5) Upsampling: This corruption arises due to interpolation errors or excessive point generation. To simulate this, certain points are selected based on severity, and new points are generated in their vicinity, with a maximum deviation of \( \ell_\infty = 0.08 \).

	By incorporating these simulated noise corruptions into experiments, algorithm robustness and performance can be comprehensively evaluated, facilitating the development of more resilient methods capable of handling real-world point cloud imperfections.

	\subsubsection{Transformation perturbations}

\begin{figure}[ht]
	\centering
	\includegraphics[width=\linewidth]{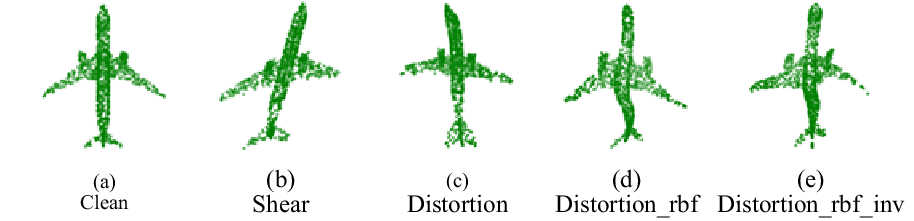}
	\caption{Visualization of transformation perturbations in point clouds.  (a) Clean point cloud. (b) Motion-induced distortion from unstable scanning. (c) Deformations mimicking structural flexibility. (d)Smooth global warping using radial basis functions. (e) Exaggerated warping effects from inverse radial basis functions transformations.}
	\label{fig:CorruptionTransformation}
	\end{figure}

	Transformation perturbations refer to geometric errors that occur during data acquisition, processing, and transmission, leading to deformation or displacement. These perturbations primarily arise from three sources: improper calibration, where inaccurate spatial positioning results from uncalibrated equipment; mechanical instability, where vibrations or unstable scanning conditions introduce spatial distortions; and processing errors, where inaccuracies or incorrect algorithm implementations during tasks such as fusion or reconstruction cause misalignment. The primary types of transformation corruption include Shear, Distortion \cite{chen2024integrated}, Distortion based on Radial Basis Function (Distortion\_rbf) \cite{de2023energy}, and Distortion based on Inverse Radial Basis Function (Distortion\_rbf\_inv). The impact of these transformations on point clouds is visually demonstrated in \cref{fig:CorruptionTransformation}.

	To systematically simulate transformation corruptions in point clouds, the following methods are applied:

	(1) Shear: This corruption represents motion distortion in point clouds, typically caused by unstable scanning. To simulate this effect, a shear transformation is applied on the plane, distorting spatial relationships \cite{yang2022lidar}.

	(2) Distortion: This corruption introduces flexible structural deformations, mimicking the effects observed in soft materials. The distortion is applied by manipulating control points in 3D space, generating localized geometric perturbations.

	(3) Distortion\_rbf: This corruption introduces smooth deformations across the entire point cloud using a radial basis function. To simulate this, five control points are placed along each axis, forming a total of 125 control points. The deformation distance is determined by severity, with randomly assigned directional shifts in 3D space. Multi-quadric radial basis functions (\(\varphi(\mathbf{x}) = \sqrt{\mathbf{x}^2 + r^2}\)) are employed to formalize the deformation.

	(4) Distortion\_rbf\_inv: This corruption applies an inverse multi-quadric radial basis function transformation, creating exaggerated warping effects. Similar to Distortion\_rbf, 125 control points are used, and deformation severity determines the extent of spatial perturbations. The inverse multi-quadric function (\(\varphi(\mathbf{x}) = \left(\mathbf{x}^2 + r^2\right)^{-\frac{1}{2}}\)) is utilized to generate the deformation.

	By incorporating these simulated transformation corruptions into experiments, the robustness of algorithms can be effectively assessed, facilitating the development of techniques capable of handling real-world geometric distortions in point clouds.

    \section{Method}
    \label{sec:Method}

    \subsection{Overview of SRRF framework}
\begin{figure}[!htbp]
		\centering
		\includegraphics[width=\linewidth]{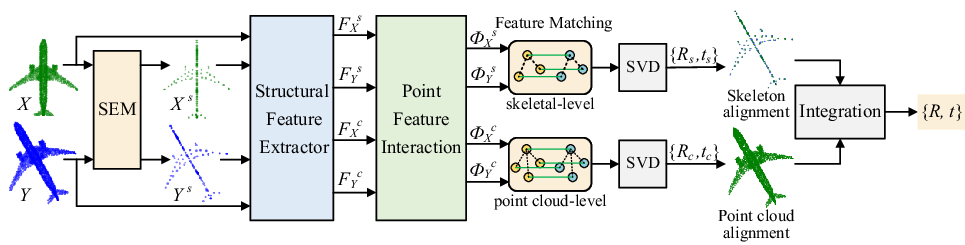}\\
		\caption{Overview of the SRRF. The framework consists of four main stages: (1) Skeletal Extraction Module, (2) Structural Feature Extraction and Interaction, (3) Feature Matching and Transformation Estimation, and (4) Transformation Integration.}
		\label{fig:PipelineSRRF}
	\end{figure}

	To effectively address the challenges posed by corrupted point clouds in registration tasks, the SRRF is proposed. This framework leverages the intrinsic skeletal structure of point clouds to enhance robustness and accuracy during alignment processes. The overall framework is illustrated in \cref{fig:PipelineSRRF}.

	The SRRF framework consists of the following key modules:

	(1) Skeletal extraction:	Given a pair of corrupted point clouds \(X\) (source) and \(Y\) (target), the Skeletal Extraction Module (SEM) extracts their respective skeletal structures, denoted as \(X^s\) and \(Y^s\). These skeletal representations serve as a compact yet informative abstraction of the original point clouds, preserving essential geometric features while reducing sensitivity to noise and density variations.

	(2) Structural feature extraction and interaction: The Structural Feature Extractor (SFE) module is responsible for extracting feature descriptors from both the original corrupted point clouds and their skeletal representations. This process generates feature sets \(F_X^c\) and \(F_Y^c\) for the corrupted point clouds, as well as \(F_X^s\) and \(F_Y^s\) for the skeletal structures. These features effectively capture both local and global geometric relationships, providing a solid foundation for robust correspondence estimation. Subsequently, the extracted feature sets are fed into the Point Feature Interaction (PFI) module, where deep feature interactions and contextual information are learned. Through this process, more discriminative features are obtained, producing \( \mathnormal{\Phi}_X^c \) and \( \mathnormal{\Phi}_Y^c \) for the corrupted point clouds, and \( \mathnormal{\Phi}_X^s \) and \( \mathnormal{\Phi}_Y^s \) for the skeletal structures. This interaction enhances the expressiveness of the features by incorporating both local and global contexts, thereby improving the reliability and accuracy of point cloud registration.

	(3) Geometric matching and transformation estimation: A Geometric Matching Module establishes a soft correspondence between the extracted discriminative features of the source and target point clouds. A differentiable SVD is then applied to estimate two transformation sets: \( \{R_s, t_s\} \) for aligning the skeletal structures, \( \{R_c, t_c\} \) for aligning the corrupted point clouds.

	(4) Transformation integration: The final step involves integrating the transformations obtained from both the corrupted point clouds and their skeletal structures to derive a global transformation \( \{R, t\} \). This integration ensures that the alignment accounts for both the detailed geometric features and the skeletal structure’s stability, leading to a robust and accurate registration outcome.

	By incorporating skeletal information and emphasizing feature interactions, the SRRF effectively mitigates the adverse effects of point cloud corruption, resulting in improved performance in registration tasks. The details of these models are given in the following subsection.

	\subsection{Skeleton extraction module}

\begin{figure}[!h]
        \centering
        \includegraphics[width=\linewidth]{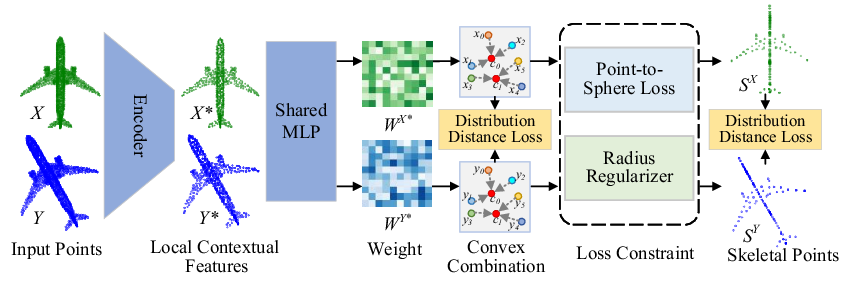}\\
        \caption{The overall pipeline of SEM. Local features are extracted from two input point clouds, processed through a shared MLP, and used in a convex combination to estimate skeletal points and radii. Structural consistency is enforced through a loss constraint, including distribution distance, point-to-sphere loss, and a radius regularizer.}
        \label{fig:Fig_SEM}
    \end{figure}

    The SEM aims to extract skeletal representations of underlying shape structures in complex point clouds. Lin et al. \cite{lin2021point2skeleton} leveraged MAT and deep neural networks to learn generalized skeletal representations. However, this framework is primarily designed for a general purpose, and might not be well-suited for point cloud registration due to the possible inconsistency between the skeleton of the source point cloud and the skeleton of the target point cloud, which consequently causes large rotation and translation errors, as shown in \cref{fig:ConceptSkeletonRegistration}(b). Inspired by this approach, a SEM tailored for point cloud registration is proposed. The pipeline is illustrated in \cref{fig:Fig_SEM}. This module extracts skeletons from both source and target point clouds simultaneously and introduces a distribution distance loss function to ensure consistency between the skeletons.

\begin{figure}[!h]
        \centering
        \includegraphics[width=\linewidth]{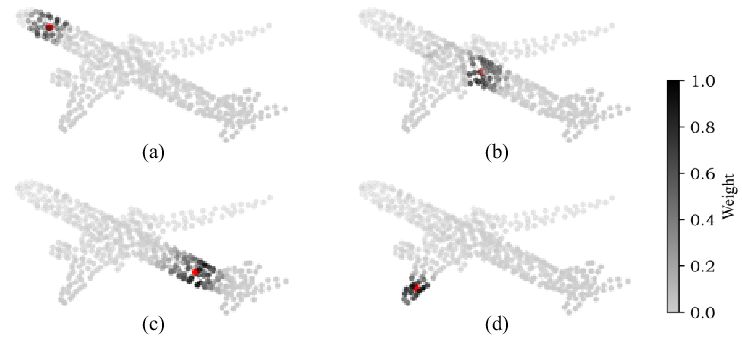}\\
        \caption{Visualization of the combined weights of skeleton points. The prediction weight of a skeleton point (marked in red) is only valid for input points within the local neighborhood of the skeleton point, while the weight gradually decreases to 0 for input points far away from the skeleton point.}
        \label{fig:ConvexCombination}
    \end{figure}

    Given point clouds $X = \left\{ {{x_1},{x_2}, \cdots ,{x_N}} \right\} \in {\mathbb{R}^{N \times 3}}$ and $Y = \left\{ {{y_1},{y_2}, \cdots ,{y_M}} \right\} \in {\mathbb{R}^{M \times 3}}$, SEM aims to extract the corresponding skeleton ${S^X} \in {\mathbb{R}^{{N_X} \times 3}}$ and ${S^Y} \in {\mathbb{R}^{{N_Y} \times 3}}$, where ${N_X}$ and $N{}_Y$ are the number of skeleton points. Taking point cloud $X$ as an example,  PointNet++ \cite{chen2024pose} is employed as an encoder to obtain the sampled points ${X^*}$ and their contextual features  ${F_{X^*}} \in {\mathbb{R}^{{N_X^*} \times D}}$, where, ${N_X^*} < N$ and $D$ is the dimension of the contextual features. A MLP and a softmax layer are used to predict the weights ${ {W}_{{X}^*}} \in {\mathbb{R}^{\left| {{X^*}} \right| \times {N_X}}}$ for the convex combinations of the sampled points. The weights are shared between ${X^*}$ and ${Y^*}$. The convex combination of the sampled points ${X^*}$ is calculated to obtain the skeleton point ${S^X}$ as:
    \begin{equation}\label{5}
        S^X = W_{X^*}^TX^* \quad s.t. \quad j = 1, \ldots ,N  \quad and \quad  \sum\limits_{i = 1}^{|{X^*}|} {{{\cal W}_{X^*}}} (i,j) = 1
    \end{equation}
    where ${W_{X^*}}$ are the combination weights and $X^*$ are the sampled points. This formula ensures that the skeleton point $S^X$ is a weighted average of the sampled points, with weights determined by the local neighborhood of $S^X$. The visualization of the weight combinations for some skeleton points is shown in \cref{fig:ConvexCombination}.

    To further optimize the skeleton, the positions of skeleton points are constrained within their local neighborhoods using the radii of skeleton spheres. The process of calculating the radius of each skeleton sphere is conducted as follows. First, the nearest distance from a sampled point \( X^* \) to all skeleton points \( \{ s_i \} \) is determined as follows:
    \begin{equation}\label{6}
        d\left(x, \{s_i\}\right)=\min_{s \in \{s_i\}} \|x-s\|_2
    \end{equation}

	Next, the distances from all input points are aggregated into a vector \( D^X \in \mathbb{R}^{|X^*| \times 1} \). The radius of each skeleton point is then calculated as a linear combination of the nearest distances from all input points, i.e., ${R^X} = {\cal W}_{X^*}^{\mathrm T}{D^X}$.

	\subsection{Structural feature extraction and interaction}
	The SFE and the PFI collectively enhance the structural representation and interaction of point cloud features. The SFE focuses on extracting both local and global structural dependencies, addressing the limitations of CNN-based Transformers in capturing contextual relationships. The PFI, in contrast, facilitates deep feature interaction between independent point cloud features, ensuring improved discriminability.

    \begin{figure}[!h]
		\centering
		\includegraphics[width=\linewidth]{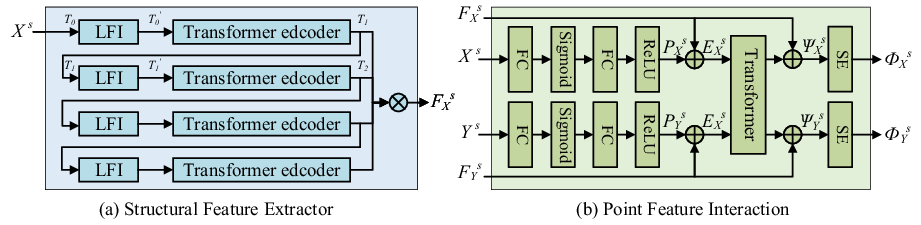}\\
		\caption{Transformer-based structural feature extraction and interaction. where \(\otimes\) denotes concatenation, \(\oplus\) denotes matrix addition.}
		\label{fig:fig_SFE+PFI}
	\end{figure}

	By modeling the structural dependencies in point clouds, SFE exhibits robustness to noise during feature extraction. \cref{fig:fig_SFE+PFI}(a) shows the architecture of SFE, which consists of Local Feature Integrator (LFI) and Transformer encoders. The LFI progressively structures the point cloud by identifying local geometric patterns through the k-nearest neighbors method. Taking the skeleton point cloud \(X^s\) as an example, the \(n\)th LFI layer identifies a local neighborhood \(P_n\) for each point, containing \(k\) nearest neighbors. The local structural information is then integrated by concatenating feature vectors from neighboring points to construct an enriched feature representation \(T_n'\):
	\begin{equation}
		\label{eq:eq_1}
		T_n' = \text{Concat}([T^1_{P_n}, T^2_{P_n}, ..., T^k_{P_n}])
	\end{equation}

	With the enriched local features, the Transformer encoder models global dependencies using Multi-Head Self-Attention (MSA) \cite{zhao2021point}. Each Transformer encoder layer consists of a MLP, Layer Normalization (LN), and MSA, which is formulated as:
	\begin{equation}
		\hat{T} = \text{LN}(\text{MSA}(T_n') + T_n')
	\end{equation}
	\begin{equation}
		T_{n+1} = \text{LN}(\text{MLP}(\hat{T}) + \hat{T})
	\end{equation}
	where \( T_{n+1} \) represents the input to the next LFI layer. MSA applies Multi-Head Attention (MA), defined as:
	\begin{equation}
		\text{MA}(Q, K, V) = \text{Concat}(A_1, \dots, A_h) W^O,
	\end{equation}
	\begin{equation}
		A_i = \text{Att}(F_Q W_i^Q, F_K W_i^K, F_V W_i^V)
	\end{equation}
	\begin{equation}
		\text{Att}(Q, K, V) = \text{softmax} \left( \frac{QK^T}{\sqrt{d_K}} \right) V.
	\end{equation}
	where \( Q \), \( K \), and \( V \) are linearly projected representations of the input \(T_n'\). \( W_i^Q, W_i^K \), and \( W_i^V \) are the projection matrices used to project \( F_Q, F_K \), and \( F_V \) into queries \( Q \), keys \( K \), and values \( V \), respectively; \( d_K \) is the dimensionality of \( K \); and \( W^O \) is a matrix used to project the concatenated features.

	Finally, the hierarchical integration of both low-order and high-order features is achieved by concatenating the Transformer encoder outputs:
	\begin{equation}
		F_X^s = \text{LN}(\text{ReLU}(\text{Concat}(F_2, F_3, \dots, F_{N_l+1}))).
	\end{equation}
	where \(F_X^s\) is the composite feature of the skeleton \( X^s \) extracted by SFE. The feature extraction process of the skeleton feature \(F_Y^s\), the corrupt point cloud feature \(F_X^c\) and \(F_Y^c\) is similar to the above.

	The features extracted by SFE have limited discriminative power because they are independent of each other. To solve this problem, PFI aims to capture contextual information and enhance feature interactions between point clouds. As shown in \cref{fig:fig_SFE+PFI}(b), PFI consists of a Transformer-based encoder-decoder and a position encoding network.

	The position encoding network learns the relative position between points. It consists of a fully connected layer (FC), ReLU and an sigmoid activation. The position encoding network extracts the position information \(P_X^s\), \(P_X^s\) as
	\begin{equation}
	\begin{aligned}
		P_X^s = \text{ReLU}(\text{FC}(\text{Sigmoid}(\text{FC}(X^s)))) \\
		P_Y^s = \text{ReLU}(\text{FC}(\text{Sigmoid}(\text{FC}(Y^s))))
	\end{aligned}
	\end{equation}

	Positional encodings \( P_X^s \) and \( P_Y^s \) are then added to \( F_X^s \) and \( F_Y^s \) to refine the features as \( E_X^s \) and \( E_Y^s \).

	To aggregate contextual information from different point clouds simultaneously, a standard Transformer \(\phi\) is adopted, which consists of an encoder and a decoder. The Transformer decoder consists of a multi-head cross-attention (MCA), in addition to MSA, MLP, and LN. Taking the feature interaction \(\Phi(E_X^s,E_Y^s)\) between skeletons as an example, the procedure of the decoder is defined as
	\begin{equation}
		\begin{aligned}
			\Psi_X^s = F_X^s + \phi(E_Y^s, E_X^s)\\
			\Psi_Y = F_Y^s + \phi(E_X^s, E_Y^s)
		\end{aligned}
	\end{equation}

	To further optimize feature importance, a Squeeze-and-Excitation (SE) module\cite{hu2018squeeze} is introduced. It first applies average pooling to generate a channel descriptor, then maps it to channel weights via a neural network. The input features are rescaled accordingly, yielding the final PFI-enhanced features are:
	\begin{equation}
		\begin{aligned}
			\Phi_X^s = \text{SE}(F_X + \phi(F_Y + P_Y, F_X + P_X))\\
			\Phi_Y^s = \text{SE}(F_Y + \phi(F_X + P_X, F_Y + P_Y))
		\end{aligned}
	\end{equation}

	%By integrating positional encoding, cross-attention, and SE-based recalibration, PFT improves feature discrimination and interaction in point cloud processing.

	\subsection{Geometric matching and transformation estimation}
	To address the limitations of non-differentiable hard assignments, a probabilistic soft-matching approach is introduced, enabling continuous and differentiable point cloud correspondences. Rather than discrete mappings, this method generates a soft map, allowing a smooth probabilistic assignment between points. Taking the registration between skeleton pairs as an example, corrupt point cloud pairs are similar. Given a skeleton point \( x_i \in X^s \), its probability distribution over elements in \( Y^s \) is formulated as:
	\begin{equation}
		m(x_i^s, Y^s) = \text{softmax}(\mathnormal{\Phi}_Y^s \mathnormal{\Phi}_{x_i^s}^{\top})
	\end{equation}
	where \( \mathnormal{\Phi}_Y^s \) represents the learned embedding of \( Y \), and \( \mathnormal{\Phi}_{x_i^s} \) denotes the \( i \)th row of \( \mathnormal{\Phi}_X^s \). This formulation facilitates soft pointer generation, where \( m(x_i^s, Y^s) \) defines a probabilistic correspondence from \( x_i^s \) to elements of \( Y^s \).

	The estimated soft correspondences are then utilized to compute the transformation estimation. The corresponding matched point \( \hat{y}_i^s \) in \( Y \) for each \( x_i^s \) is determined as:

	\begin{equation}
		\hat{y}_i^s = Y^{\top} m(x_i^s, Y^s) \in \mathbb{R}^{3},
	\end{equation}
	where \( Y^s \in \mathbb{R}^{N \times 3} \) is the matrix containing the points in \( Y^s \). The transformation estimation \( \{ R_s, t_s \} \) between skeleton pairs are then estimated based on the paired correspondences \( x_i^s \rightarrow \hat{y}_i^s \). To ensure differentiability, SVD is applied, providing an efficient and stable computation for optimizing the rigid transformation. Following the same principle, the transformation estimation \( \{R_c, t_s \} \) between pairs of corrupt point clouds can be calculated.

	\subsection{Transformation integration}
	Finally, To achieve a robust and reliable transformation estimation in rigid point cloud registration, it is necessary to integrate two distinct transformation estimates derived from different alignment strategies: one obtained from the corrupted point cloud alignment \( \{R_c, t_c\} \) and the other from the skeleton-based alignment \( \{R_s, t_s\} \). Since each estimation captures complementary structural characteristics, a weighted fusion approach is employed to optimally combine their contributions. The final transformation estimation \( R, t \) is formulated as a weighted combination of both estimates:
	\begin{equation}
		\{R, t\} = \lambda \{R_c, t_c\} + (1-\lambda) \{R_s, t_s\}
	\end{equation}
	where \( \lambda \in [0,1] \) is an adaptive weighting factor that regulates the relative influence of each transformation estimate. The inlier ratio \( \lambda \) reflects the proportion of correspondences that are geometrically consistent, which can be calculated as:
	\begin{equation}
		\lambda = \frac{\gamma_c}{\gamma_c + \gamma_s}
	\end{equation}
	\begin{equation}
		\gamma_c = \frac{N_c^{\text{inlier}}}{N_c}
	\end{equation}
	\begin{equation}
		\gamma_s = \frac{N_s^{\text{inlier}}}{N_s}
	\end{equation}
	where \( \gamma_c \) is the inlier ratio for the corrupted point cloud alignment. \( \gamma_s \) is the inlier ratio for the skeleton-based alignment. \( N_c^{\text{inlier}} \) and \( N_s^{\text{inlier}} \) are the number of inliers detected for each transformation. \( N_c \) and \( N_s \) are the total number of correspondences.

	This formulation ensures that the transformation estimate with a higher inlier ratio is assigned a greater weight. By integrating these two transformation estimates, this fusion approach capitalizes on the robustness of the skeleton-based alignment, which effectively captures global structural consistency, while incorporating information from the corrupted point cloud registration, which retains local geometric details. Consequently, the proposed method enhances the accuracy, stability, and reliability of the final transformation estimation in rigid point cloud registration.

    \subsection{Loss function}
    \label{sec:Loss Function}
    A registration loss and a skeleton loss are used to supervise the proposed framework. The registration loss consists of two terms that quantify the discrepancy between the predicted and ground truth rotation and translation components.

     The registration loss \( L_{reg} \) is calculated as follows:
    \begin{equation}
    	{L_{reg}} = {\left\| {{R^{\mathrm T} }{R_{gt}} - I} \right\|^2} + {\left\| {t - {t_{gt}}} \right\|^2}
    \end{equation}
    where \(gt\) means the ground truth, \(I\) represents the identity matrix, and the notation \( \left\| \cdot \right\|^2 \) represents the squared norm of a matrix or vector.

    The skeleton loss is used to supervise the SEM, which consists \(L_{bsp}\) and \(L_{ddl}\). The loss \(L_{bsp}\) is used to generate basic skeleton points, which is the weighted sum of the sampling loss ${L_{\rm{s}}}$, point-to-sphere loss ${L_p}$, and radius regularization loss ${L_r}$\cite{lin2021point2skeleton}.
    \begin{equation}
        {L_{bsp}} = {L_s} + {\lambda _1}{L_p} + {\lambda _2}{L_r}
    \end{equation}
    where $\lambda _1$ and $\lambda _2$ are hyperparameters balancing the losses.

	To further enhance the consistency of distribution between the source and target skeletons, a distribution distance loss, denoted as \( L_{ddl} \), is designed. This loss function is specifically formulated to reduce the discrepancy between the transformed source skeleton and the target skeleton. The \( L_{ddl} \) between the source skeleton \( X^s \) and the target skeleton \( Y^s \) is defined as:
    \begin{equation}
    	L_{ddl} = \sum\limits_{x^s\in X^s}\min\limits_{y^s\in Y^s}\left\|x^s-y^s\right\|_2 + \sum\limits_{y^s\in Y^s}\min\limits_{x^s\in X^s}\left\|y^s-x^s\right\|_2
    \end{equation}

	To achieve a more stable training process and improved performance, the gradient flow from the skeleton extraction module to the skeleton registration module is prevented. Through this comprehensive approach, both the registration accuracy and the integrity of the skeletal structure are ensured.

    \section{Experiments and discussion}
    \label{sec:Experiments and discussion}
	In \cref{sec:Datasets and Experimental Setup}, the datasets, implementation details, and evaluation metrics used in this study are outlined. \cref{sec:Comparison of registration performance under different types of corruption} provides a comparison of point cloud registration performance under density corruption, noise corruption, and transform corruption to validate the effectiveness and robustness of the current method. Finally, \cref{sec:Ablation Study} presents ablation studies, verifying the effectiveness of point cloud simplification levels, simplification methods, and the distribution distance loss function.

    \subsection{Datasets and experimental setup}
    \label{sec:Datasets and Experimental Setup}
    \subsubsection{Dataset}
    ModelNet40 \cite{modelnet40} is a widely used dataset for point cloud registration, comprising 12,308 point clouds distributed across 40 categories. These point clouds are extracted from CAD models, resulting in a perfectly clean dataset. However, in real-world scenarios, point cloud corruption is inevitable due to scene complexity, sensor inaccuracies, and processing errors. Models trained on clean datasets often perform poorly when tested on corrupted point clouds. To evaluate the registration performance of various models in corrupt point clouds, a corrupt point cloud dataset was constructed using the simulated corruption method described in \cref{sec:Corruption point cloud dataset construction}.

    \subsubsection{Experimental setup}
    the proposed model is implemented in PyTorch, and all experiments are conducted on a GeForce RTX 2080 GPU. The Adam optimizer is used to optimize the network parameters, with an initial learning rate of 0.001. The learning rate is reduced by a factor of 10 at epochs 75, 150, and 200, and the training runs for a total of 250 epochs.

    \subsubsection{Metrics}
    The model is evaluated using Mean Squared Error (MSE), Root Mean Squared Error (RMSE), and Mean Absolute Error (MAE) between the ground truth and the predicted values of the rotation matrix and translation vector. Ideally, these error metrics should be zero if the rigid alignment is perfect. All angular measurements in experimental results are reported in degrees.

    \subsection{Registration experiments under different corruption types}
    \label{sec:Comparison of registration performance under different types of corruption}
    Corrupted point clouds represent one of the most challenging scenarios in point cloud registration, a critical issue frequently encountered in real-world applications.
    To comprehensively evaluate the performance of the proposed method under corrupted point clouds, comparative experiments were conducted across three common corruption types: density variation, noise contamination, and transformation perturbation. The current method is systematically compared with six mainstream approaches (ICP \cite{vizzo2023kiss}, PointNetLK \cite{aoki2019pointnetlk}, DCP \cite{wang2019deep}, RGM \cite{fu2021robust}, PANet \cite{wu2023panet}, and Equi-GSPR \cite{kang2024equi}) to assess its robustness and effectiveness.
    \subsubsection{Registration performance under different density variation}
	Density corruption in point cloud data can significantly impact the performance of registration algorithms. The considered types of density corruption include density increase, density decrease, and local cropping, which effectively simulate the characteristics of point cloud data generated by various real-world sensing systems. To assess the robustness of the proposed method under different density corruption scenarios, a comparative analysis was conducted against benchmark algorithms. The registration performance of the current method and the competing methods under these conditions is presented in \cref{tab:tab_1}.
	\begin{table}[!h]
      	\centering
		\caption{Registration performance under different density variation.}
		\label{tab:tab_1}
		\resizebox{\linewidth}{!}{%
		\begin{tabular}{llllllll}
			\hline
			Corruption               & Model         & \multicolumn{1}{c}{MSE(R)} & \multicolumn{1}{c}{RMSE(R)} & \multicolumn{1}{c}{MAE(R)} & \multicolumn{1}{c}{MSE(t)} & \multicolumn{1}{c}{RMSE(t)} & \multicolumn{1}{c}{MAE(t)} \\ \hline
			\multirow{7}{*}{Density\_inc} & ICP           & 127.043625                 & 11.271363                   & 12.259451                  & 0.071226                   & 0.266881                    & 0.084291                   \\
			& PointNetLK    & 79.357633                  & 8.908290                    & 8.619321                   & 0.044356                   & 0.210609                    & 0.056165                   \\
			& DCP           & 47.522358                  & 6.893646                    & 4.465775                   & 0.005951                   & 0.077140                    & 0.018610                   \\
			& RGM           & 32.565155                  & 5.706589                    & 4.015452                   & 0.004466                   & 0.066825                    & 0.014152                   \\
			& PANet         & 16.464545                  & 4.057653                    & 3.984525                   & 0.001546                   & 0.039325                    & 0.010145                   \\
			& Equi-GSPR     & 10.486485                  & 3.238284                    & 3.904545                   & 0.000695                   & 0.026354                    & 0.006847                   \\
			& \textbf{Ours} & \textbf{7.456156}          & \textbf{2.730596}           & \textbf{2.651815}          & \textbf{0.000459}          & \textbf{0.021419}           & \textbf{0.005515}          \\ \hline
			\multirow{7}{*}{Density\_dec} & ICP           & 127.224892                 & 11.279401                   & 13.216220                  & 0.068427                   & 0.261585                    & 0.088680                   \\
			& PointNetLK    & 72.156253                  & 8.494484                    & 9.652354                   & 0.041651                   & 0.204086                    & 0.062654                   \\
			& DCP           & 55.869884                  & 7.474616                    & 5.252159                   & 0.002176                   & 0.046647                    & 0.036107                   \\
			& RGM           & 30.165152                  & 5.492281                    & 5.016516                   & 0.002057                   & 0.045349                    & 0.037647                   \\
			& PANet         & 19.565465                  & 4.423287                    & 4.654545                   & 0.001655                   & 0.040687                    & 0.011646                   \\
			& Equi-GSPR     & 10.354655                  & 3.217865                    & 3.883451                   & 0.001155                   & 0.033980                    & 0.006715                   \\
			& \textbf{Ours} & \textbf{7.848185}          & \textbf{2.801461}           & \textbf{2.815165}          & \textbf{0.000866}          & \textbf{0.029420}           & \textbf{0.005684}          \\ \hline
			\multirow{7}{*}{Cutout}       & ICP           & 144.308075                 & 12.012830                   & 12.926389                  & 0.088265                   & 0.297095                    & 0.130321                   \\
			& PointNetLK    & 78.651163                  & 8.868549                    & 9.822165                   & 0.034651                   & 0.186148                    & 0.093541                   \\
			& DCP           & 52.713661                  & 7.260417                    & 5.301618                   & 0.001995                   & 0.044665                    & 0.034828                   \\
			& RGM           & 39.564562                  & 6.290037                    & 5.012655                   & 0.001655                   & 0.040677                    & 0.016556                   \\
			& PANet         & 21.564646                  & 4.643775                    & 4.241646                   & 0.001265                   & 0.035573                    & 0.007565                   \\
			& Equi-GSPR     & 11.412456                  & 3.378233                    & 3.764516                   & \textbf{0.000825}          & \textbf{0.028715}           & \textbf{0.006445}          \\
			& \textbf{Ours} & \textbf{8.181426}          & \textbf{2.860319}           & \textbf{3.298456}          & 0.000873                   & 0.029547                    & 0.006816                   \\ \hline
		\end{tabular}
		}
	\end{table}

	The experimental results presented in \cref{tab:tab_1} demonstrate that the current method exhibits strong robustness across different density corruption scenarios. The skeletal structure effectively accommodates both local density increases and decreases, ensuring high registration accuracy under varying conditions. Compared to mainstream methods, the current method consistently achieves lower errors, particularly in rotation estimation. Under cutout corruption, the model successfully identifies the correct rotation direction but exhibits a slightly higher translation error due to missing regions where skeletal points cannot be extracted. Overall, the results confirm that the current method effectively extracts and utilizes skeletal structures, significantly enhancing registration accuracy in point cloud data with local density variations and demonstrating strong resilience to density transformations.

    \subsubsection{Registration performance under different noise contamination}
    In this section, the registration performance of the current method is evaluated under various noise contamination scenarios. The tested noise types include uniform noise, Gaussian noise, impulse noise, and background noise, which collectively simulate common disturbances introduced by sensor inaccuracies and data preprocessing artifacts. These noise models reflect real-world challenges encountered in point cloud acquisition. The experimental results, summarized in \cref{tab:tab_2}.
	\begin{table}[!h]
		\centering
		\caption{Registration performance under different noise contamination.}
		\label{tab:tab_2}
		\resizebox{\linewidth}{!}{%
		\begin{tabular}{ccllllll}
			\hline
			Corruption            & Method        & \multicolumn{1}{c}{MSE(R)} & \multicolumn{1}{c}{RMSE(R)} & \multicolumn{1}{c}{MAE(R)} & \multicolumn{1}{c}{MSE(t)} & \multicolumn{1}{c}{RMSE(t)} & \multicolumn{1}{c}{MAE(t)} \\ \hline
			\multirow{7}{*}{Uniform}    & ICP           & 120.580666                 & 10.980923                   & 11.817094                  & 0.089305                   & 0.298840                    & 0.255872                   \\
			& PointNetLK    & 69.562462                  & 8.340411                    & 6.542823                   & 0.088165                   & 0.296925                    & 0.304562                   \\
			& DCP           & 34.050190                  & 5.835254                    & 3.544258                   & 0.084509                   & 0.290704                    & 0.250847                   \\
			& RGM           & 29.545615                  & 5.435588                    & 3.432556                   & 0.052157                   & 0.228378                    & 0.190256                   \\
			& PANet         & 12.056456                  & 3.472241                    & 3.304565                   & 0.006647                   & 0.081528                    & 0.156542                   \\
			& Equi-GSPR     & 9.856456                   & 3.139499                    & 3.254615                   & 0.001023                   & 0.031987                    & 0.074155                   \\
			& \textbf{Ours} & \textbf{6.956262}          & \textbf{2.637473}           & \textbf{2.900556}          & \textbf{0.000864}          & \textbf{0.029396}           & \textbf{0.054612}          \\ \hline
			\multirow{7}{*}{Gaussian}   & ICP           & 128.448227                 & 11.333500                   & 12.407916                  & 0.090786                   & 0.038447                    & 0.224536                   \\
			& PointNetLK    & 71.625303                  & 8.463173                    & 6.865432                   & 0.041606                   & 0.203974                    & 0.101519                   \\
			& DCP           & 56.343520                  & 7.506232                    & 4.778652                   & 0.006026                   & 0.077625                    & 0.017009                   \\
			& RGM           & 41.464455                  & 6.439290                    & 4.155616                   & 0.006002                   & 0.077470                    & 0.016855                   \\
			& PANet         & 19.556546                  & 4.422278                    & 3.698655                   & 0.000593                   & 0.024360                    & 0.010016                   \\
			& Equi-GSPR     & 9.956432                   & 3.155382                    & 3.378462                   & 0.000584                   & 0.024175                    & 0.006256                   \\
			& \textbf{Ours} & \textbf{8.054145}          & \textbf{2.837983}           & \textbf{3.144652}          & \textbf{0.000467}          & \textbf{0.021601}           & \textbf{0.005616}          \\ \hline
			\multirow{7}{*}{Impulse}    & ICP           & 134.627213                 & 11.602897                   & 12.504189                  & 0.094493                   & 0.307397                    & 0.276202                   \\
			& PointNetLK    & 62.562562                  & 7.909650                    & 6.854626                   & 0.090547                   & 0.300909                    & 0.295462                   \\
			& DCP           & 39.797520                  & 6.308527                    & 3.615429                   & 0.087867                   & 0.296424                    & 0.256603                   \\
			& RGM           & 28.256256                  & 5.315661                    & 3.446946                   & 0.065466                   & 0.255862                    & 0.256246                   \\
			& PANet         & 13.343453                  & 3.652869                    & 3.392355                   & 0.022453                   & 0.149845                    & 0.103414                   \\
			& Equi-GSPR     & 10.433846                  & 3.230146                    & 3.362456                   & 0.009668                   & 0.098328                    & 0.074151                   \\
			& \textbf{Ours} & \textbf{7.946515}          & \textbf{2.818956}           & \textbf{3.044652}          & \textbf{0.005652}          & \textbf{0.075177}           & \textbf{0.065626}          \\ \hline
			\multirow{7}{*}{Background} & ICP           & 363.250580                 & 19.059133                   & 16.333392                  & 0.122450                   & 0.349929                    & 0.160874                   \\
			& PointNetLK    & 112.265256                 & 10.595530                   & 9.625265                   & 0.051645                   & 0.227256                    & 0.124515                   \\
			& DCP           & 52.033530                  & 7.213427                    & 5.263993                   & 0.006348                   & 0.079674                    & 0.061400                   \\
			& RGM           & 31.256513                  & 5.590752                    & 5.116566                   & 0.005985                   & 0.077360                    & 0.059546                   \\
			& PANet         & 19.456465                  & 4.410948                    & 4.935646                   & 0.003656                   & 0.060469                    & 0.036545                   \\
			& Equi-GSPR     & \textbf{12.956456}         & \textbf{3.599508}           & \textbf{4.307855}          & 0.001865                   & 0.043191                    & 0.010165                   \\
			& \textbf{Ours} & 13.496119                  & 3.673706                    & 4.454612                   & \textbf{0.000182}          & \textbf{0.013475}           & \textbf{0.008856}          \\ \hline
		\end{tabular}
	     }
	\end{table}

	The experimental results presented in \cref{tab:tab_2} demonstrate that the current method effectively handles various noise contamination scenarios, including uniform, Gaussian, impulse, and background noise. The strong performance of the current method can be attributed to its ability to extract and utilize robust structural features, minimizing the influence of noise-induced distortions. Specifically, under uniform and Gaussian noise, where disturbances are either evenly distributed or randomly fluctuate around the true values, the current method maintains stable registration accuracy. This is due to its effective feature extraction, which helps filter out small-scale variations while preserving essential geometric structures. In the presence of impulse noise, which introduces sporadic extreme outliers, the current method significantly reduces the impact of these outliers by prioritizing reliable skeletal features over isolated erroneous points. This makes it particularly robust in scenarios where sensor data is prone to occasional large deviations. For background noise, which adds extraneous points that do not belong to the original structure, the current model effectively differentiates between relevant and irrelevant data. While some competing methods struggle with misalignment due to the presence of additional noise points, the current method maintains high registration accuracy by focusing on meaningful structural cues. Overall, the results validate that the current method is highly robust to different types of noise, effectively suppressing outliers and irrelevant variations while maintaining precise registration. This adaptability makes it well-suited for real-world applications where point cloud data is often subject to sensor noise and preprocessing artifacts.

    \subsubsection{Registration performance under different transformation perturbation}
    Unlike density and noise corruption, which primarily affect local features while preserving the overall geometric structure, transformation perturbations introduce global or local deformations that significantly alter both the position and shape of point clouds. These non-rigid transformations, including shearing, free-form deformation, and radial basis function (RBF) deformation, present greater challenges for point cloud registration as they disrupt spatial consistency and structural integrity. To evaluate the robustness of the current method in handling such complex transformations, experiments were conducted under these perturbation conditions. Shearing distortions introduce non-uniform transformations that affect the relative spatial arrangement of points, making it difficult for traditional methods to establish accurate correspondences. Free-form deformations further complicate the registration process by introducing irregular warping, often encountered in dynamic data acquisition scenarios. Similarly, RBF-based deformations simulate nonlinear distortions common in augmented and virtual reality applications, as well as generative model outputs, which require high adaptability from registration algorithms. The experimental results are presented in \cref{tab:tab_3}.
	\begin{table}[!h]
        \centering
		\caption{Registration performance under different transformation perturbation.}
		\label{tab:tab_3}
		\resizebox{\linewidth}{!}{%
		\begin{tabular}{ccllllll}
			\hline
			Corruption                      & Method        & \multicolumn{1}{c}{MSE(R)} & \multicolumn{1}{c}{RMSE(R)} & \multicolumn{1}{c}{MAE(R)} & \multicolumn{1}{c}{MSE(t)} & \multicolumn{1}{c}{RMSE(t)} & \multicolumn{1}{c}{MAE(t)} \\ \hline
			\multirow{7}{*}{Shear}                & ICP           & 246.091904                 & 15.687316                   & 15.154194                  & 0.109038                   & 0.330210                    & 0.256627                   \\
			& PointNetLK    & 99.321629                  & 9.966024                    & 9.518287                   & 0.098722                   & 0.314200                    & 0.257846                   \\
			& DCP           & 86.778400                  & 9.315493                    & 7.008987                   & 0.087273                   & 0.295420                    & 0.254685                   \\
			& RGM           & 49.654846                  & 7.046619                    & 6.875154                   & 0.083555                   & 0.289058                    & 0.254111                   \\
			& PANet         & 23.659565                  & 4.864110                    & 5.013255                   & 0.035646                   & 0.188801                    & 0.102355                   \\
			& Equi-GSPR     & 12.068782                  & 3.474015                    & 4.781526                   & 0.012465                   & 0.111649                    & 0.086545                   \\
			& \textbf{Ours} & \textbf{11.056165}         & \textbf{3.325081}           & \textbf{4.254597}          & \textbf{0.009102}          & \textbf{0.095402}           & \textbf{0.066123}          \\ \hline
			\multirow{7}{*}{Distortion}           & ICP           & 161.590942                 & 12.711842                   & 10.228945                  & 0.122532                   & 0.350045                    & 0.273413                   \\
			& PointNetLK    & 92.625261                  & 9.624202                    & 8.015626                   & 0.098432                   & 0.313739                    & 0.264615                   \\
			& DCP           & 79.322330                  & 8.906308                    & 6.665538                   & 0.072393                   & 0.269059                    & 0.226312                   \\
			& RGM           & 58.565453                  & 7.652807                    & 5.982555                   & 0.068985                   & 0.262650                    & 0.215456                   \\
			& PANet         & 30.156646                  & 5.491507                    & 5.015646                   & 0.039322                   & 0.198297                    & 0.110547                   \\
			& Equi-GSPR     & 10.517256                  & 3.243032                    & 4.764525                   & 0.015456                   & 0.124321                    & 0.078546                   \\
			& \textbf{Ours} & \textbf{8.951206}          & \textbf{2.991857}           & \textbf{3.845152}          & \textbf{0.004566}          & \textbf{0.067574}           & \textbf{0.061853}          \\ \hline
			\multirow{7}{*}{Distortion\_rbf}      & ICP           & 163.161697                 & 12.773476                   & 11.309789                  & 0.092747                   & 0.304544                    & 0.259079                   \\
			& PointNetLK    & 65.628216                  & 8.101124                    & 6.554922                   & 0.089045                   & 0.298404                    & 0.269456                   \\
			& DCP           & 39.810630                  & 6.309566                    & 4.384098                   & 0.084597                   & 0.290855                    & 0.250852                   \\
			& RGM           & 27.645656                  & 5.257914                    & 4.164555                   & 0.051412                   & 0.226742                    & 0.204589                   \\
			& PANet         & 16.565565                  & 4.070082                    & 4.013547                   & 0.013555                   & 0.116424                    & 0.143626                   \\
			& Equi-GSPR     & 9.063455                   & 3.010557                    & 3.989645                   & 0.008535                   & 0.092388                    & 0.126897                   \\
			& \textbf{Ours} & \textbf{8.432965}          & \textbf{2.903957}           & \textbf{3.231646}          & \textbf{0.004146}          & \textbf{0.064386}           & \textbf{0.068462}          \\ \hline
			\multirow{7}{*}{Distortion\_rbf\_inv} & ICP           & 154.192444                 & 12.417425                   & 10.141214                  & 0.092419                   & 0.304005                    & 0.258753                   \\
			& PointNetLK    & 69.612580                  & 8.343415                    & 8.562856                   & 0.046185                   & 0.214906                    & 0.105652                   \\
			& DCP           & 45.290500                  & 6.729822                    & 4.600007                   & 0.003634                   & 0.060279                    & 0.044787                   \\
			& RGM           & 30.455246                  & 5.518627                    & 4.005889                   & 0.002156                   & 0.046435                    & 0.018426                   \\
			& PANet         & 18.564646                  & 4.308671                    & 3.953162                   & 0.001989                   & 0.044594                    & 0.011656                   \\
			& Equi-GSPR     & 9.941556                   & 3.153023                    & 3.564566                   & 0.001470                   & 0.038338                    & 0.009646                   \\
			& \textbf{Ours} & \textbf{9.198246}          & \textbf{3.032861}           & \textbf{3.129565}          & \textbf{0.000621}          & \textbf{0.024928}           & \textbf{0.005146}          \\ \hline
		\end{tabular}
		}
	\end{table}

	The experimental results presented in \cref{tab:tab_3} demonstrate the robustness of the current method in handling various transformation perturbations. These perturbations introduce significant structural alterations to the point cloud, posing a challenge to traditional and deep learning-based registration methods. Specifically, Under shearing transformations, the current method achieves superior registration accuracy compared to mainstream approaches. Shearing alters the relative spatial arrangement of points, often leading to misalignment in rigid registration techniques. The current method effectively mitigates these effects by leveraging robust feature extraction, resulting in lower rotation and translation errors. For free-form deformations, which introduce complex, non-uniform warping, traditional methods such as ICP and even deep learning-based models struggle to maintain registration accuracy. The current method significantly reduces errors by effectively capturing geometric structures and maintaining alignment despite severe distortions. This highlights its adaptability to real-world dynamic data acquisition scenarios.  In the case of distortion\_rbf deformations, which simulate non-rigid transformations encountered in generative models and augmented reality applications, the current method consistently achieves the lowest registration errors. This indicates its ability to handle highly nonlinear transformations where many competing methods fail due to their reliance on rigid or semi-rigid constraints.

	 Overall, the results validate that the current method is highly resilient to different types of transformation perturbations. The improved performance can be attributed to the ability of the current method to adaptively extract and utilize high-level geometric representations, ensuring more reliable registration even when traditional rigid constraints fail. By analyzing these results, valuable insights into the impact of non-rigid corruption on registration accuracy are gained, and potential strategies to enhance robustness in complex real-world scenarios are identified.

    \subsection{Ablation study}
    \label{sec:Ablation Study}
    \subsubsection{Effectiveness of skeleton-based sampling in point cloud registration}
\begin{figure}[!h]
    	\centering
    	\includegraphics[width=\linewidth]{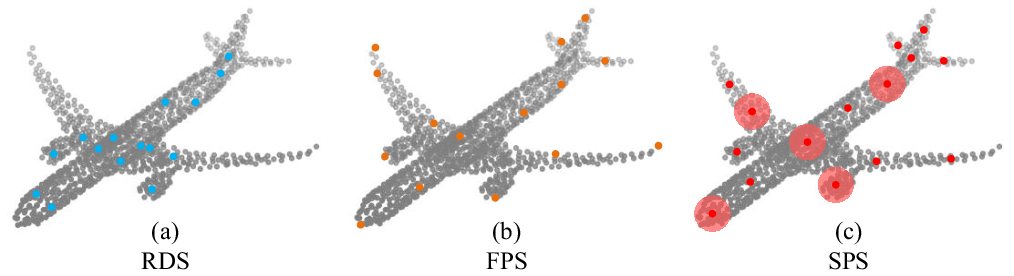}
    	\caption{Visualization of point clouds using different simplification methods. (a) random downsampling, (b) farthest point sampling, (c) skeleton point sampling. The small red dots represent the sampled skeleton points, while the large red circles represent the range of original points covered by the corresponding skeleton points obtained.}
    	\label{fig:ThreeSampling}
    \end{figure}

    To evaluate the influence of different point cloud simplification methods on registration accuracy, three representative sampling strategies, Random Downsampling (RDS), Farthest Point Sampling (FPS), and Skeleton Point Sampling (SPS), were examined. RDS is a straightforward approach that randomly selects a subset of points, though it risks overlooking essential structural features. FPS, in contrast, selects points farthest from those already chosen, preserving better geometric representativeness. SPS, however, focuses on selecting structural key points that are crucial for describing the overall shape of the point cloud, thereby retaining fundamental geometric information. \cref{fig:ThreeSampling} illustrates these three sampling strategies, highlighting their respective characteristics in point cloud simplification.

\begin{figure}[!ht]
        \centering
        \subfloat[]{
            \includegraphics[scale=0.43]{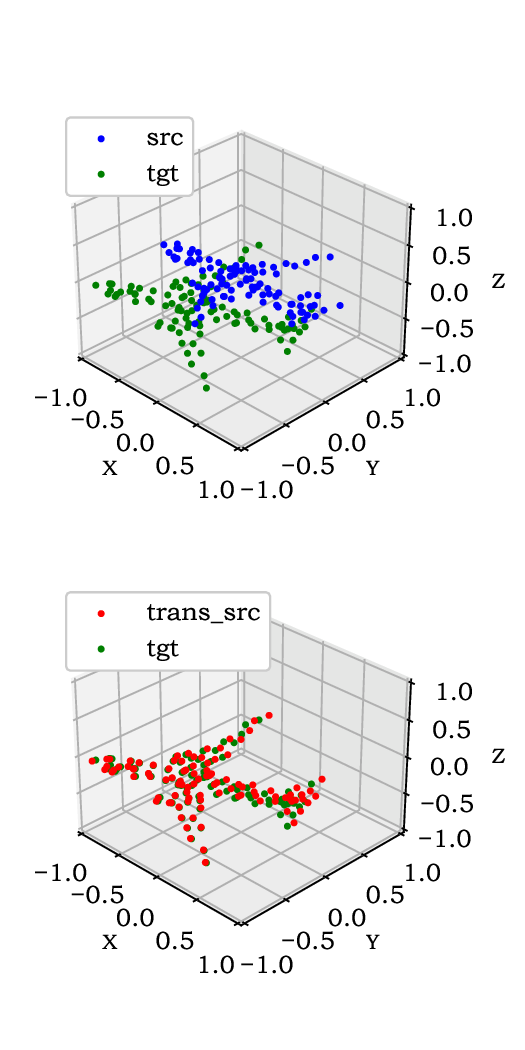}}
        \subfloat[]{
            \includegraphics[scale=0.43]{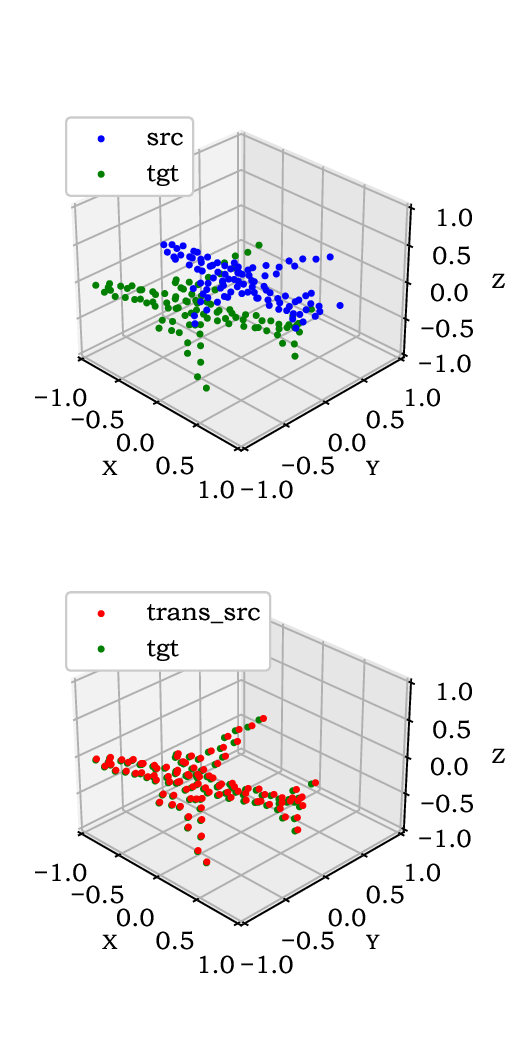}}
        \subfloat[]{
            \includegraphics[scale=0.43]{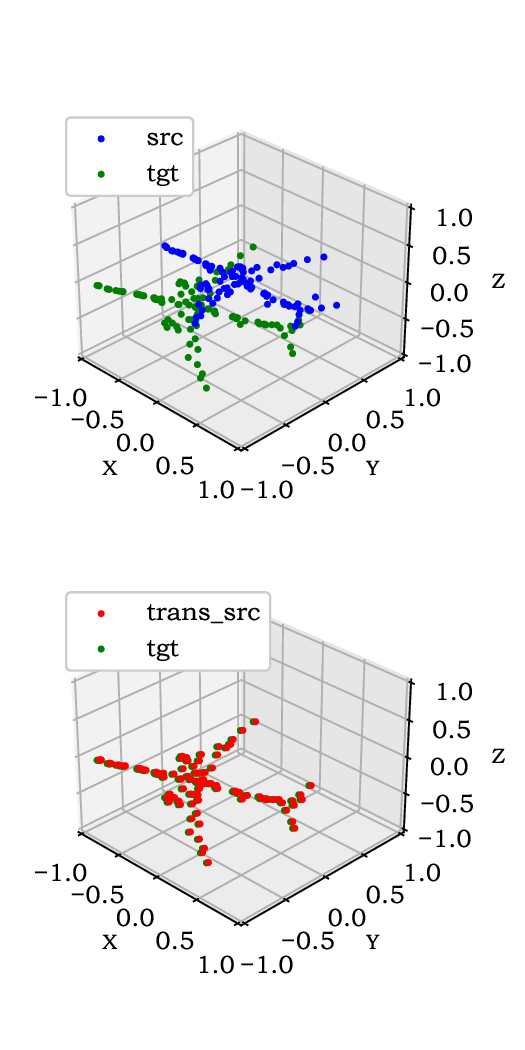}}
        \caption{Registration visualization under different simplification strategies. The source skeleton is marked in blue, the target skeleton is marked in green, and the transformed source skeleton is marked in red. (a) random sampling. (b) farthest point sampling. (c) skeleton point sampling. The structural representation of skeleton points is more compact and representative, resulting in better registration results.}
        \label{fig:ThreeSamplingRegistration}
    \end{figure}

    To quantitatively analyze the impact of these simplification strategies on registration accuracy, registration experiments were conducted using the classical deep learning-based registration algorithm DCP \cite{wang2019deep}. The registration performance under each sampling strategy is presented in \cref{tab:tab_4}, and the corresponding visualization is shown in \cref{fig:ThreeSamplingRegistration}. The experimental results indicate that RDS does not significantly reduce the registration error. This is because the sampling points are arbitrarily selected, and it is difficult to ensure that there is a relatively accurate correspondence between the sampled point clouds to be aligned. This lack of structural consistency will lead to misalignment and reduced registration accuracy. FPS improves upon RDS by prioritizing spatial coverage and ensuring the selection of points that represent the global shape of the point cloud. However, due to the inherent randomness in selecting the initial points, FPS cannot fully guarantee that the downsampled source and target point clouds will maintain perfect structural correspondence, leading to potential mismatches during registration. In contrast, SPS consistently outperforms all other methods across all evaluation metrics, particularly in rotational error. This highlights its superior ability to capture the topological structure of point clouds. Unlike RDS and FPS, which may introduce inconsistencies in correspondence between sampled source and target point clouds, SPS offers a more robust solution by integrating local and global structural information. By focusing on crucial skeletal points, this method ensures high similarity between the downsampled source and target point clouds, leading to significantly improved registration accuracy. Overall, the results demonstrate the importance of selecting points that capture fundamental structural information for high-precision point cloud registration. The strong performance of SPS suggests that leveraging skeletal structures enhances robustness, making it particularly suitable for complex transformations and real-world applications requiring reliable and accurate alignment.

        \begin{table}[!ht]
   	\centering
   	\caption{Registration performance under different simplification methods.}
   	\label{tab:tab_4}
   	\resizebox{\linewidth}{!}{%
   		\begin{tabular}{ccccccc}
   			\toprule
   			Method & R\_MSE & R\_RMSE & R\_MAE & t\_MSE & t\_RMSE & t\_MAE \\
   			\midrule
   			Original & 6.571834 & 2.563559 & 1.513092 & 0.000003 & 0.001735 & 0.001483 \\
   			RDS & 4.861980 & 2.204990 & 1.370186 & 0.000003 & 0.001715 & 0.001032 \\
   			FPS & 5.276511 & 2.297066 & 1.287942 & 0.000003 & 0.00186 & 0.001046 \\
   			SPS & 2.144033 & 1.464252 & 0.887999 & 0.000001 & 0.001129 & 0.000617\\
   			\bottomrule
   		\end{tabular}
   	}
   \end{table}

    \subsubsection{Effectiveness of distance loss in the distribution of skeleton points}
    In the process of skeleton point generation, a distribution distance loss function was designed to improve the distribution consistency between the source and the target skeleton. This experiment aims to evaluate the distance between the matching points of the transformed source skeleton and the target skeleton, so as to prove the effectiveness of the loss function. The evaluation metrics is the chamfer distance \(d_{CD}\), which is formalized as:
    \begin{equation}\label{10}
        d_{\mathrm{CD}}(S_1,S_2)={10^{ - 4}} \times(\sum\limits_{x\in S_1}\min\limits_{y\in S_2}\lVert x-y\rVert_2^2+\sum\limits_{y\in S_2}\min\limits_{x\in S_1}\lVert y-x\rVert_2^2)
    \end{equation}
    where $S_1$ and $S_2$ represent the sets of skeleton points from the source and target point clouds, respectively.

    \begin{table}[!h]
        \centering
        \caption{The Effectiveness of the Distribution Distance Loss Function.}
        \label{tab:tab_5}
        %\resizebox{\linewidth}{!}{%
            \begin{tabular}{cc}
            \toprule
                \multicolumn{1}{c}{\hspace{1cm} Method} & \hspace{2cm} $d_{\mathrm{CD}}$ \hspace{1cm} \\
            \midrule
                \multicolumn{1}{c}{\hspace{1cm} w/o $L_d$} & \hspace{2cm} 3.85 \hspace{1cm} \\
                \multicolumn{1}{c}{\hspace{1cm} w/ $L_d$} & \hspace{2cm} 1.83 \hspace{1cm} \\
            \bottomrule
            \end{tabular}
        %}
    \end{table}

    \begin{figure}[!htbp]
        \centering
        \subfloat{
            \includegraphics[scale=0.45]{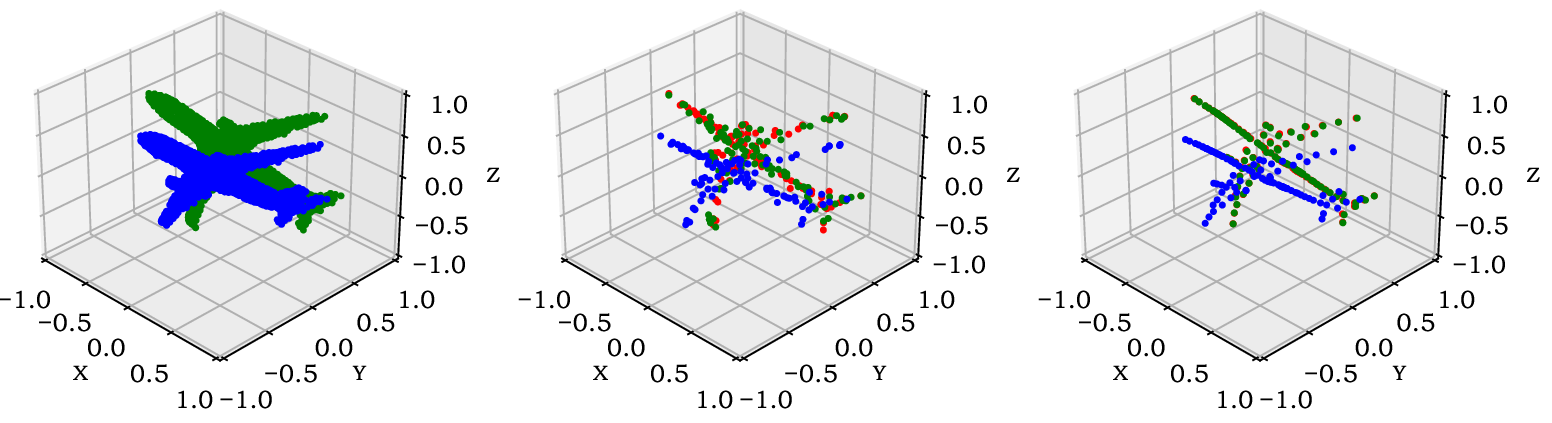}}\\
        \subfloat{
            \includegraphics[scale=0.45]{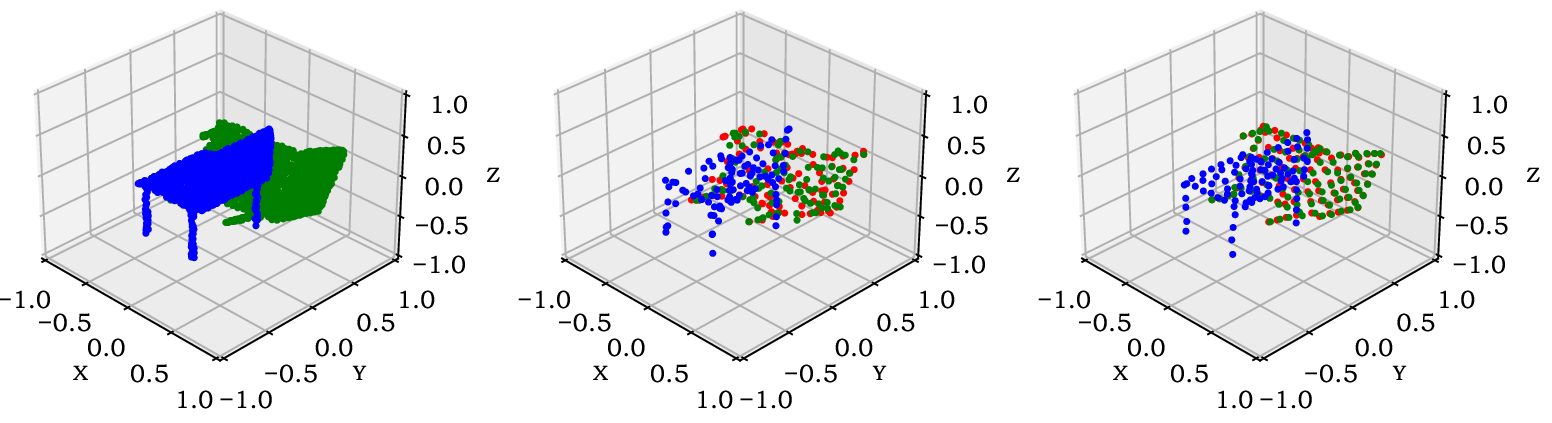}}\\
        \subfloat{
            \includegraphics[scale=0.45]{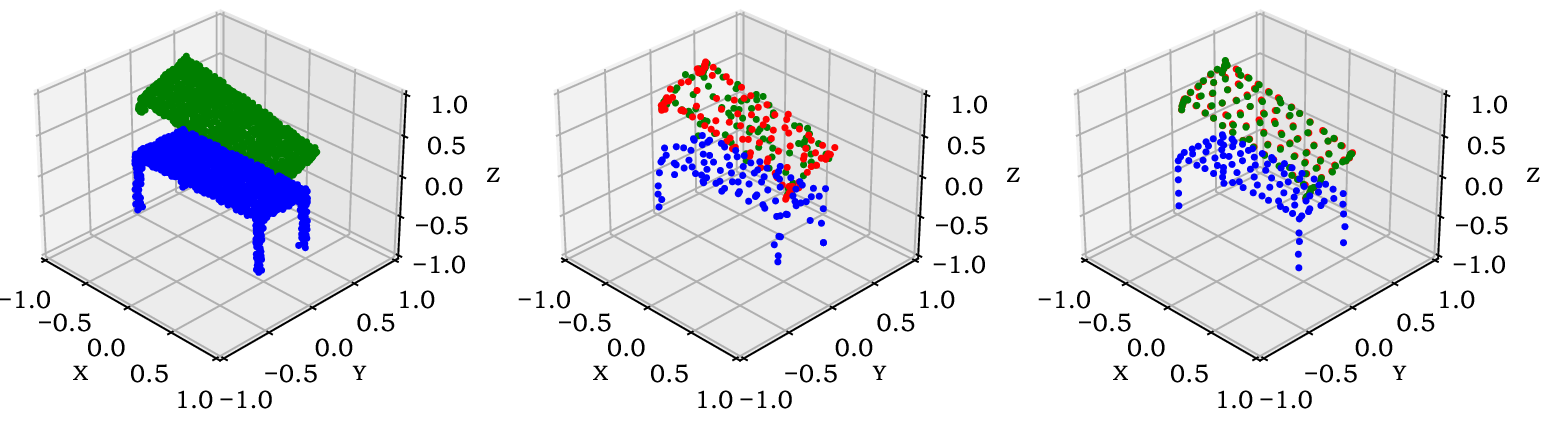}}\\
        \subfloat{
            \includegraphics[scale=0.45]{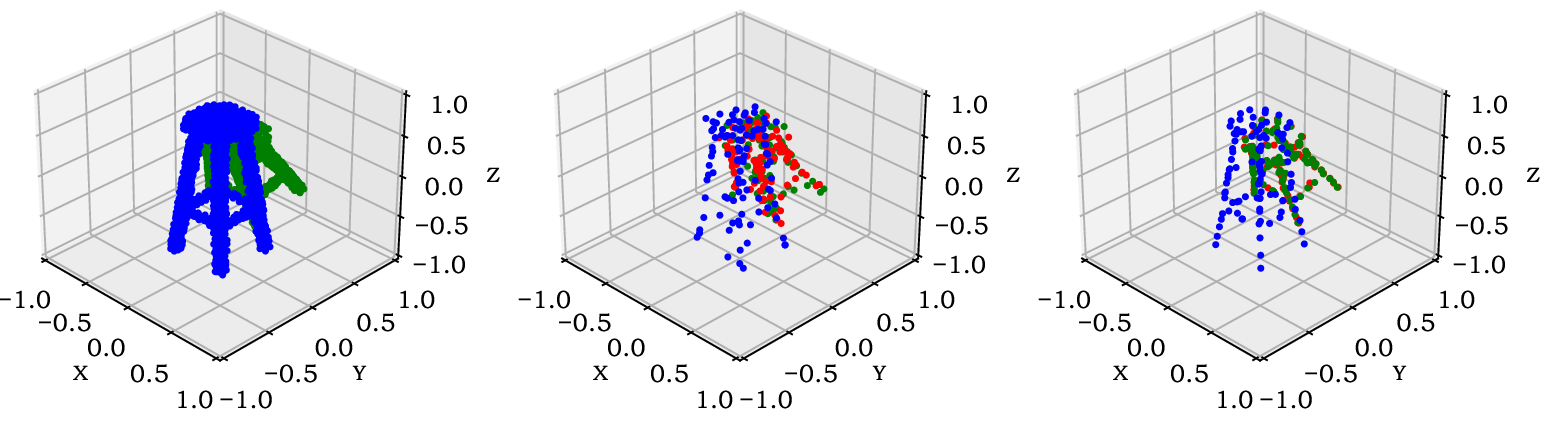}}\\
        \subfloat{
            \includegraphics[scale=0.45]{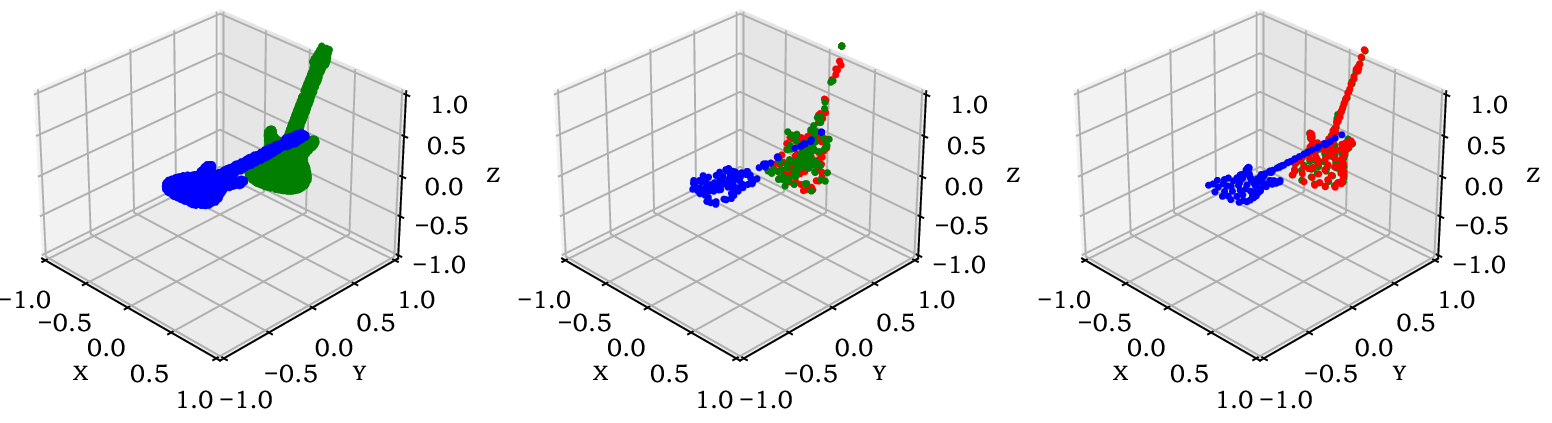}}\\
        \caption{Ablation studies on distribution distance loss. The leftmost figures represent the original point clouds, while the middle and rightmost figures depict the extracted skeletons without and with the loss function, respectively. In these figures, blue and green represent the original source and target point clouds, while red denotes the transformed source skeleton after applying the ground truth transformation.}
        \label{fig:LossSkeleton}
    \end{figure}

    The quantitative experiments are shown in \cref{tab:tab_5}. The experimental results show that after the loss function is introduced, the distribution distance between the pairs of point clouds to be registered is significantly reduced, indicating that the pairs of skeletons to be registered can be better aligned. This confirms that the loss function effectively optimizes the matching degree between the skeleton pairs, which is conducive to more accurate registration.

	To more intuitively demonstrate the effectiveness of the distance loss function, the skeleton extraction results before and after applying the loss function are shown in \cref{fig:LossSkeleton}. The visualization results indicate that incorporating the distribution distance loss leads to a more consistent structure and a better-aligned skeleton representation, which contributes to enhancing the robustness of registration in the presence of noise and disturbances.

    \section{Conclusion}
    \label{sec:Conclusion}
    In this study, the SRRF was proposed to address the challenges posed by corrupted point cloud data. Unlike conventional methods that rely on raw point distributions and are highly sensitive to noise, density variations, and geometric deformations, SRRF leverages skeleton structures to provide a stable and informative representation for registration. By integrating transformations obtained from both the corrupted point cloud alignment and its skeleton alignment, SRRF ensures a robust and accurate registration process that considers both local geometric details and global structural stability. Extensive experimental evaluations across diverse corruption scenarios demonstrated that SRRF consistently outperforms existing approaches, validating its effectiveness for real-world applications.

    Despite its advantages, the current method has some limitations. while SRRF demonstrates strong resilience to common corruptions, it may face challenges when handling extreme non-rigid deformations, which require more advanced deformation modeling. Integrating learning-based deformation estimation into SRRF could further improve its adaptability. In addition, our current implementation assumes a relatively uniform level of corruption across the dataset. In practical applications, point cloud corruptions may be highly localized and non-uniform. Future work could explore adaptive weighting strategies to handle heterogeneous corruptions more effectively.

    Overall, SRRF lays a solid foundation for robust point cloud registration in real-world conditions and opens new avenues for advancing corruption-aware 3D vision applications.

	\section*{List of Abbreviations}
	\begin{table}[H]
		\centering
		\renewcommand{\arraystretch}{1.2}
%		\caption{List of abbreviations used in this paper.}
		\begin{tabular}{ l  l }
			\hline
			\textbf{Abbreviation} & \textbf{Full Form} \\
			\hline
			FC    & Fully Connected \\
			FPS   & farthest point sampling \\
			KNN   & K-Nearest Neighbors  \\
			LFI   & Local Feature Integrator \\
			LN    & Layer Normalization \\
			MAE   & Mean Absolute Error \\
			MAT   & Medial Axis Transformation \\
			MCA   & Multi-head Cross-Attention \\
			MLP   & Multi-Layer Perceptron \\
			MSA   & Multi-Head Self-Attention \\
			MSE   & Mean Squared Error \\
			PFI   & Point Feature Interaction \\
			RDS   & Random Downsampling \\
			RMSE  & Root Mean Squared Error \\
			SE    &  Squeeze-and-Excitation \\
			SEM   & Skeletal Extraction Module \\
			SFE   & Structural Feature Extractor \\
			SPS   & Skeleton Point Sampling \\
			SVD   & Singular Value Decomposition \\
			SRRF  & Skeleton-Based Robust Registration Framework \\
			\hline
		\end{tabular}
		\label{tab:abbreviations}
	\end{table}

    \section*{Acknowledgment}
    This work was partially supported by the National Natural Science Foundation of China under Grants (No.51774219), and the key R\&D Program of Hubei Province (No.2020BAB098).

    Numerical calculation is supported by High-Performance Computing Center of Wuhan University of Science and Technology.

    % \bibliographystyle{unsrt}
    % \bibliography{refs}

\end{document}